\documentclass[runningheads]{IEEEtran}
\usepackage{hyperref}
\usepackage{amsfonts,mathtools}
\usepackage{color,xcolor}
\usepackage{array,flushend}
\usepackage{tabularx,caption}
\usepackage{rotating,multirow,booktabs}
\usepackage{amsmath,amssymb,amsfonts,bm,mwe}
\usepackage{graphicx,subcaption,subfloat}
\usepackage{tikz}
\usepackage[skip=2pt,font=small]{caption}
\usetikzlibrary{positioning}

\begin{document}
%

\title{
Improving Neural Networks for Time-Series Forecasting using Data Augmentation and AutoML}
%
%
\author{\IEEEauthorblockN{Indrajeet Y. Javeri, Mohammadhossein Toutiaee, Ismailcem B. Arpinar, and John A. Miller}\\ \IEEEauthorblockA{Department of Computer Science\\
University of Georgia, Athens, Georgia\\
Email: \{iyj83163, mt31429, budak, jamill\}@uga.edu}\\ 
\IEEEauthorblockA{Tom W. Miller\\
Department of Economics, Finance \& Quantitative Analysis\\
Kennesaw State University, Kennesaw, Georgia\\
Email: tmiller@kennesaw.edu}}

%
%
%
\maketitle              

\begin{abstract}
Statistical methods such as the Box-Jenkins method for time-series forecasting have been prominent since their development in 1970. Many researchers rely on such models as they can be efficiently estimated and also provide interpretability. However, advances in machine learning research indicate that neural networks can be powerful data modeling techniques, as they can give higher accuracy for a plethora of learning problems and datasets. In the past, they have been tried on time-series forecasting as well, but their overall results have not been significantly better than the statistical models especially for intermediate length times series data. Their modeling capacities are limited in cases where enough data may not be available to estimate the large number of parameters that these non-linear models require. This paper presents an easy to implement data augmentation method to significantly improve the performance of such networks. Our method, Augmented-Neural-Network,  which involves using forecasts from statistical models, can help unlock the power of neural networks on intermediate length time-series and produces competitive results. It shows that data augmentation, when paired with Automated Machine Learning techniques such as Neural Architecture Search, can help to find the best neural architecture for a given time-series. Using the combination of these, demonstrates significant enhancement in the forecasting accuracy of three neural network-based models for a COVID-19 dataset, with a maximum improvement in forecasting accuracy by 21.41\%, 24.29\%, and 16.42\%, respectively, over the neural networks that do not use augmented data.


\end{abstract}

\section{Introduction}
In recent years, neural networks have been used extensively in time-series forecasting to capture the nonlinear information that traditional linear statistical models cannot fully capture, and where nonlinear regression models may be challenging to develop.
Earlier models also tried combining neural networks with traditional models where linear components of the time-series data were captured using the statistical models and the neural networks were then trained to capture the leftover non-linearities. 
While neural networks were not competitive in many earlier time-series forecasting competitions,
an automated neural network model was judged the 9th best model in the M3 time-series forecasting competition \cite{makridakis2000m3} held in 2000.
More recently, several neural network models have been a part of the M4 in 2018 and M5 in 2020 time-series forecasting competitions \cite{makridakis2018m4}, \cite{makridakis2020m5} and have been competitive and, in some cases, better than the traditional statistical models. 

A major challenge with the neural network-based models arises when we are looking at scenarios where less training data is available such that the networks are not fully able to learn the information from the limited number of samples. Short and intermediate length time-series forecasting is one of such areas where neural networks struggle to compete with other models. However, the performance of neural networks increases significantly as we get to longer length time-series as indicated in multiple studies, e.g., \cite{peng2018forecasting, peng2019knowledge, qin2017dual}.


This paper focuses on intermediate length time-series data, those with a length comparable to the M4 competition average of 1035 data points, as neural networks often dominate for longer time-series and tend to be inferior for shorter time-series. Several open datasets that record daily records concerning the COVID-19 Pandemic fall into this intermediate length category. In addition, the highly dynamic nature of pandemic data makes it challenging for time-series forecasting.
Finally, the importance of having accurate forecasts for the pandemic cannot be overestimated.

The paper provides a comparative analysis between traditional statistical models and neural network models.
In particular, it shows that the performance of regular neural networks is competitive but not better than statistical models.
However, we show that data augmentation can significantly improve the quality of out-of-sample forecasts making them better than the traditional models.
\footnote{Codebase: \url{https://github.com/indrajeet3333/covid-19-forecasting}}

We also focus on building the neural network architecture through Automated Machine Learning (AutoML) by using Neural Architecture Search (NAS) to efficiently search for the best performing model architecture for the given learning task and dataset. This combined with the data augmentation gives the neural networks large performance boosts that help them to perform better than the statistical models.

Much of the earlier work in time-series forecasting using neural networks specifies the use of time-series features from the statistical model forecasts as inputs for the neural networks. This includes the top-performing models in M4 competition such as the hybrid method of exponential smoothing and recurrent neural networks \cite{smyl2020hybrid} and the FFORMA \cite{montero2020fforma}. Although these methods do not aim to increase the sample size of the training data made available to the neural networks, they aid in providing more information to the neural networks helping them to better model the non-linear patterns within the data and thus, improving the generalization performance of these models. The data augmentation approach that we use interleaves the forecasts collected from a statistical model to build a time-series that is double the length of the original time-series providing more samples for the neural networks to train on.

Producing accurate forecasts for the COVID-19 Pandemic has been challenging due to the length of the time-series data that are available.
Only now is a year's worth of data available for the United States.
This paper examines the COVID-19 modeling studies that have been conducted and applies common modeling techniques from two main categories:
(1) statistical models, and
(2) machine learning models.
The first category of models represents a robust and relatively easy to apply set of models that should be used in modeling studies (at least as a point of comparison).
The second category, that of machine learning, supports more intense pattern matching that has the potential to produce more accurate forecasts.
The disadvantages include the necessity for more data, tuning of hyper-parameters, architecture search, and interpretability.

Modeling studies are important for several reasons.
Forecasting is important, so people and their governments know what to expect in the next few weeks. 
The COVID-19 Pandemic is unique in the sense that the last pandemic with this (or more) significance was the 1918 H1N1 Influenza Pandemic
\cite{morens20101918}.


Modeling and forecasting studies for COVID-19 are grouped into four major approaches:
a) SEIR/SIR models,
b) Simulation/Agent-based models, 
c) Curve-fitting models, and
d) Predictive models.

\noindent \textbf{SEIR/SIR} \cite{lopez2020modified}
models are common epidemiological modeling techniques that predict the pandemic course through the system of equations that divide an estimated population into different compartments: susceptible, exposed, infected, and recovered.
A set of mathematical rules govern the equations as to how the assumptions about the disease process, Non-Pharmaceutical Interventions, social mixing, and public vaccinations would affect the population.

\noindent \textbf{Simulation/Agent-based} \cite{shamil2020agent}
models simulate a community where individuals represent ``agents'' in that community.
The pandemic course spreads among the agents based upon the virus assumptions and rules about social mixing, governmental policy, and other behavioral patterns in the studied community.

\noindent \textbf{Curve-fitting} \cite{zhao2020well}
models refer to fitting a mathematical model by considering the current state and estimate the relationship between features of the pandemic.
The future of the pandemic course can be approximated fully from assumptions about the contagion, public health policies, or other limitations and uncertainties of spread disease parameters and/or experiences in other communities.

\noindent \textbf{Predictive}
models play a key role in modern pandemic prediction.
Scientists apply different statistical and machine learning algorithms for modeling the future of morbidity and mortality of the disease, and they produce state-of-the-art results even in the presence of many hidden variables. 

The rest of this paper is organized as follows:
Section 2 focuses on the COVID-19 datasets.
Related work is reviewed in Section 3. 
Section 4 describes the epidemiological dynamics.
Section 5 and 6 discuss the statistical and machine learning models.
The methodology used for data augmentation, AutoML, and rolling validation is discussed in Section 7. Finally, results are provided in Section 8, and conclusions and future work are given in Section 9.

\section{COVID-19 Datasets}

Out of numerous data sources available for tracking the COVID-19 pandemic, here in the US, we have two primary data sources which collect the information from the public health authorities throughout the country. First, \href{https://coronavirus.jhu.edu/about}{The Johns Hopkins Coronavirus Resource Center (CRC)} is a continuously updated source of COVID-19 data have their data repository stored at 
\url{https://github.com/CSSEGISandData/COVID-19}. Secondly, \href{https://covidtracking.com/about}{The COVID Tracking Project} is a volunteer organization collecting COVID-19 data used by multiple organizations and research groups. 

This paper utilizes the national COVID-19 dataset composed of data collected and stored at
\url{https://covidtracking.com/data}.
The dataset consists of COVID-19 case data for testing, hospitalization, and patient outcomes from all over the United States.

\subsection{COVID-19 Dataset - United States}

The {\sc ScalaTion} COVID-19 dataset
\url{https://github.com/scalation/data/tree/master/COVID} contains data about COVID-19 cases, hospitalizations, deaths and several more columns in the United States.
This dataset is in a CSV file with 420 rows and
18 columns.

This study focuses on the incident deaths or daily death increase column using it as a time-series for the neural networks and statistical models.

The time-series includes daily death counts in the United States starting from January 13, 2020 to March 7, 2021.
We eliminate the first 44 day records and feed our models time-series data starting February 26, 2020 when the first deaths due to COVID-19 in the United States were recorded.
After eliminating the first 44 days, the new size of the time-series is 376 days.

The COVID Tracking project has ended their data collection phase on 7th of March 2021 \cite{datacollection}. Thus, we ensure that we take the entirety of their dataset for our analysis. Ending the dataset on March 7, 2021 has the modeling advantage that vaccination effects may be ignored, but after this time, models should take vaccinations under consideration.
Unfortunately, there will initially be limited data available on vaccinations.

Other resources for further data, as pointed out by the COVID Tracking project, include the data from Centers for Disease Control and Prevention (CDC) \cite{cdcdata} and National Center for Health Statistics (NCHS) \cite{nhcsdata}.

\subsection{Data Preprocessing}
Data is preprocessed for smoothing the outliers that lie $3.5$ standard deviations away from the rolling mean. 
The rolling mean is the $6$ time points local average such that we consider 3 time-points before and after the current point.
Observations within $3.5$ standard deviations are left unchanged while the observations which are considered as outliers are replaced by the Kalman Smoothing using a Local Level Model \cite{commandeur2007introduction}.

\subsection{Optimum Lag Selection}



\begin{figure}[htbp]
\centerline{\includegraphics[width=88mm]{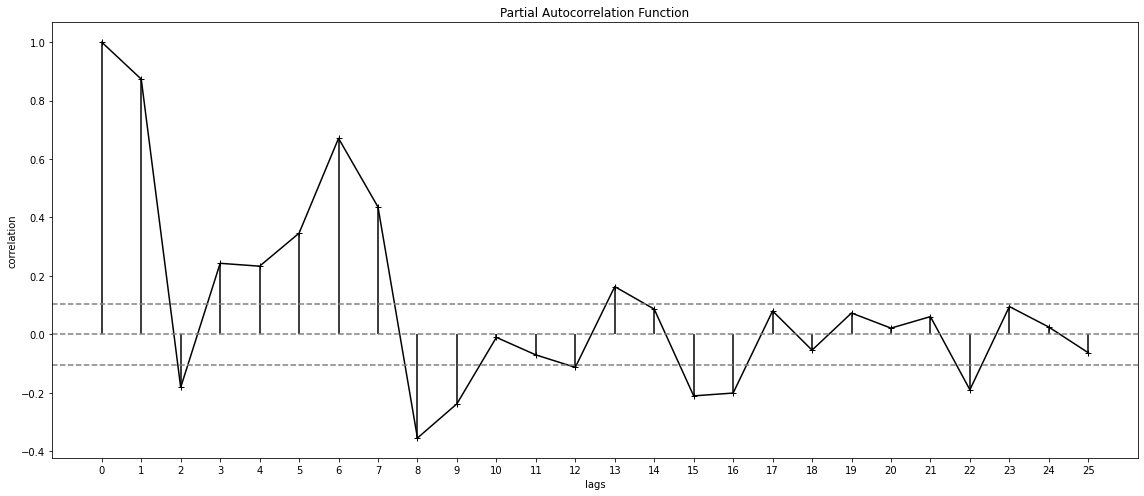}}
\caption{PACF Plot of time-series}
\label{fig:pacf}
\end{figure}

In time-series modeling, it is common to use lagged features as inputs to the models. A lag value is the target value from any of its previous values. For example, the $k^{th}$ lag value of $y_{t}$ will be  $y_{t-k}$. We need to define models such that they look for $p$ previous lags while estimating their parameters. The choice of $p$ needs to be determined before taking any model under consideration. Depending on the characteristics and dynamics of the model as well as the time-series, we choose a particular value for $p$.
We determine the optimal number of lags to be 22 using the Partial Auto-correlation function (PACF), as shown in Figure \ref{fig:pacf}.
Since, the PACF eliminates the effect of correlation with all earlier lags, significant correlations at lag 22 indicate we can use upto 22 lags for our modeling and also eliminate any in between the lags which are not significant contributors.
\begin{figure}[htbp]
\centerline{\includegraphics[width=88mm]{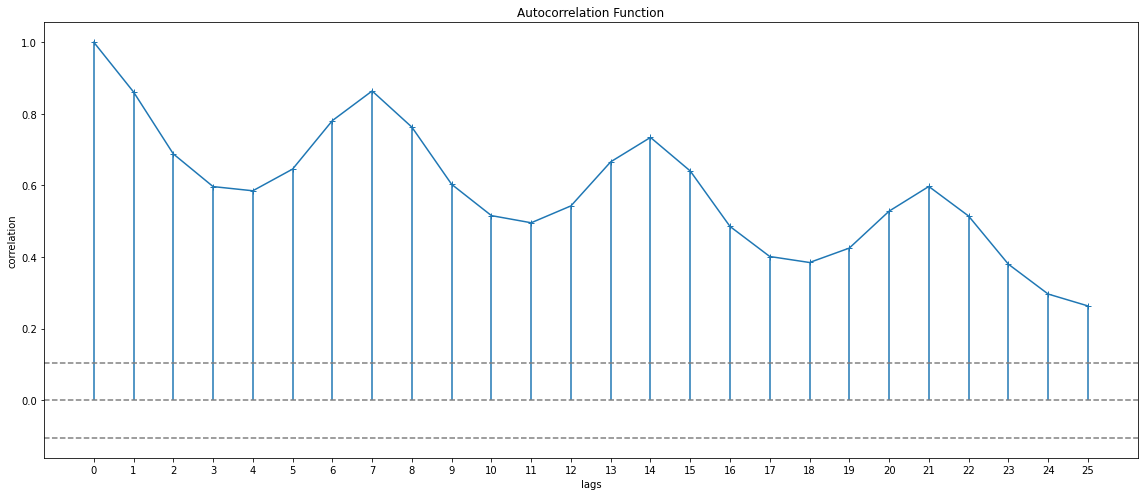}}
\caption{ACF Plot of time-series}
\label{fig:acf}
\end{figure}

\noindent

\section{Related Work}
In time-series forecasting, neural networks have been competing with many statistical models for many years. The M3 competition in the year 2000, saw a single neural network architecture being compared with the other 23 statistical models \cite{makridakis2000m3}. The Automated ANN model presented in this competition \cite{balkin2000automatic} first demonstrated the potential of using an automated procedure for the selection of neural network architecture paving way for AutoML techniques such as NAS to be used in the context of time-series forecasting. However, in the case of shorter length time-series the automatic architecture selection, even though competitive, was not better than the other statistical models considered.

In the continuation of previous competitions the M4 time-series competition, which consisted of many short and intermediate length time-series, saw a greater number of neural network-based models some of which were able to perform better than statistical models. One of the major findings of this competition included the ``hybrid" approach which included both statistical and machine learning features \cite{makridakis2018m4, makridakis2020m4}. The hybrid method by Slawek Smyl \cite{smyl2020hybrid} used a Dynamic Computational Graph Neural Network system to combine an Exponential Smoothing (statistical model) with Long Short-Term Memory (LSTM) networks to give a single hybrid and powerful forecasting method that was the winner of the M4 competition, producing forecasts which were nearly 10\% better (in sMAPE) than the combination benchmarks considered in the competition. The second most accurate method, FFORMA \cite{montero2020fforma}, also involved a combination of 7 statistical methods and 1 ML method, and the weights for averaging/combining such methods also used an ML algorithm. This encouraged the time-series research community to develop techniques that use a combination of machine learning models with time-series features as indicated in \cite{makridakis2018m4}. In light of these previous works on intermediate length time-series, we explore the performance of neural networks, compare them with statistical benchmarks and empirically show that using data augmentation can help boost the performance of such neural network models. 


Beyond Fully Connected Neural Networks, many Recurrent Neural Network (RNN) based frameworks have been proposed for intermediate length time-series forecasting as indicated in \cite{rangapuram2018deep,wang2019deep}. More recent studies such as \cite{hewamalage2021recurrent}, show the application of RNNs on several short and intermediate length time-series where the results are better than statistical models for some datasets while the RNNs performing worse than ARIMA and ETS models on diverse datasets such as the M4, which consists of 48,000 time-series. 

Convolutional Neural Networks (CNN) have also been referred to in \cite{borovykh2017conditional} which uses the dilated convolutions that can provide broader access to historical values. 
They claimed that CNN performs on par with the recurrent-type network, whilst requiring significantly fewer parameters and computation to achieve comparable performance.

With the advent of ``Attention Mechanisms'' in recent years, these techniques allow the network to directly concentrate on important time points in the past. Application of such networks in longer length time-series shows improvements over RNN-based competitors which has been reflected in  \cite{fan2019multi,li2019enhancing,lim2019temporal}. These type of networks often employs an encoder to summarize historical information and a decoder to integrate them with known future values \cite{lim2019temporal,fan2019multi}. However, there has been little evidence for the performance of such networks on shorter and intermediate length time-series data.


Although the advanced neural network architectures have achieved state-of-the-art in image and text domains, a limited number of studies are available around intermediate time-series analysis because  that such techniques require a plethora of input data for training.
Our work adds to this growing body of machine learning research with a novel approach to framing, evaluating, and comparing intermediate time-series forecasts for the COVID-19 prediction task, and in comparison with other powerful techniques, we showed that our new proposed technique (AUG-NN) can indeed outperform other competitors.

\subsection{Modeling Studies for COVID-19}

Several papers include forecasts for COVID-19 deaths, positive cases, and hospitalizations. The authors in \cite{kumar2020covid} forecasted the spread of the disease in 10 largely affected countries including the United States.
They applied ARIMA and Prophet time-series forecasting models on the number of positive cases and deaths, and they showed that ARIMA performed better than FBProphet \cite{taylor2018forecasting} on a scale of different error metrics such as MAPE and RMSE.
Another similar study that applied an ARIMA is found in \cite{ilie2020forecasting} where they extensively tested and tuned different parameters to minimize the prediction error, and they discussed that ARIMA models are appropriate for generating forecasts for the pandemic crisis.

The predictive approach by machine learning was studied by \cite{prasanth2020forecasting} where they applied the GWO-LSTM model to forecast the daily new cases, cumulative cases, and deaths for India, the USA, and the UK.
The network parameters of the LSTM network are optimized by Grey Wolf Optimizer (GWO), and they claimed that their hybrid architecture obtained a better performance compared to the baseline (ARIMA). However, they rely on access to online big data from Google Trends to evaluate other real-world conditions, in order to feed more information to the LSTMs.

\section{Statistical Time-Series Models}

The statistical time-series models considered in this work do not distinguish between state and measurement variables.
They also consider longer lags beyond, e.g., $d_{t-1}$.
In particular, four time-series models are applied for forecasting the number of deaths per day $d_t$ with a horizon of 1 to 14 days.
\textit{Random Walk (RW)} and \textit{Auto-Regressive (AR)} models are the most simple methods for solving the univariate time-series forecasting techniques that we applied in this work.
In an effort to capture the previous shocks, AR is coupled with \textit{Moving Average (MA)} component. 
This extra component, along with the \textit{Integrated (I)} component in ARIMA specifies the level of differencing for making the time-series stationary, can facilitate the discovery of more complex patterns in data.
Our preliminary experiment found usefulness in using ARIMA since we found ARIMA$(3,1,2)$ as the best global model with the lowest Akaike Information Criterion (AIC) criteria when evaluated on the training set.
Later we describe how the parameters of ARIMA can be dynamically calibrated for different horizons.
Additionally, we expect to improve the prediction accuracy by adding the \textit{Seasonality} component in ARIMA, as Figure~\ref{fig:acf} testifies such effects exist in the data at every 7 lags.
We are interested in making comparisons between our proposed approach and current statistical methods through running multiple experiments and showing that the AUG-NN technique can indeed produce the best results.
A brief overview of the statistical methods mentioned above are provided in the following:

\begin{enumerate}

\item
The \textbf{Random Walk (RW)} model states that the future value $y_t$ is the previous value $y_{t-1}$
disturbed by white noise $\epsilon_t \in N(0, I\sigma^2)$ and the forecast is just $y_{t-1}$.

\begin{equation*}
y_t ~=~ y_{t-1} + \epsilon_t
\end{equation*}

\item
The \textbf{Auto-Regressive (AR)} of Order $p$, AR$(p)$ model has
$p$ autoregressive terms (lags $[y_{t-1}, \dots, y_{t-p}]$),

\begin{equation*}
y_t = \delta + \phi_1 y_{t-1} + \dots + \phi_p y_{t-p} + \epsilon_t
\end{equation*}

where $\delta$ is the offset and $\phi_j$ is the parameter multiplying the $j^{th}$ lag, $y_{t-j}$.

\item

An \textbf{Auto-Regressive Integrated Moving Average} or ARIMA$(p,d,q)$, which is a generalization of an Auto-Regressive Moving Average (ARMA) class, provides a better prediction of future points in series, where data show evidence of non-stationarity in the mean.
The new hyper-parameter $d$ is the number of times the time-series is differenced. 
When there is a trend in the data (e.g., the values or levels are increasing over time), an ARIMA$(p, 1, q)$ may be effective. 
This model will take a first difference of the values in the time-series, namely as:
\begin{equation*}
y'_t = y_t - y_{t-1}
\end{equation*}

\item
The \textbf{Seasonal, Auto-Regressive, Integrated, Moving-Average} SARIMA$(p, d, q)_{\times}(P, D, Q)_s$ model has $p$ autoregressive terms (lags $[y_{t-1}, \dots, y_{t-p}]$), $d$ differences, $q$ moving average terms (base on errors/shocks).
In addition, it has $P$ seasonal autoregressive terms (lags $[y_{t-s}, \dots, y_{t-Ps}]$), $D$ seasonal differences, $Q$ seasonal moving average terms.
The period (or season) is given by $s$.







\end{enumerate}

\section{Machine Learning Models}

\subsection{Neural Networks}

Neural Networks allow us to capture the non-linear patterns in the data through (1) the use of multiple hidden layers, and (2) several neurons with non-linear activation functions in each hidden layer. With enough hidden layers and neurons to capture the non-linearity and enough training samples to estimate the parameters through backpropagation, neural network-based models serve as powerful data modeling techniques in the machine learning domain.
However, for pure neural network or machine learning-based models
the main disadvantage remains the inability to train a model containing a large number of parameters on intermediate length time-series.
For COVID-19 prediction with a year's worth of daily data, the length of the time-series is still shorter than ideal for training neural networks.
Therefore, in this paper, we only consider some simpler network architectures, rely on AutoML to find better models, and utilize data augmentation to compensate.

A simple type of Neural Network generalizes an AR$(p)$ model. 
We use a three-layer (two hidden) neural network, where the input layer has a node for each of the $p$ lags and the current time $t$. The hidden layer has $n_h$ nodes while the output layer has $n_o$.

\begin{eqnarray*}
\mathbf{x}_t & = & [y_{t-p},\ldots,y_{t-1}, t]\\
\mathbf{y_t} &=& W_3^t\,\mathbf{f_2}(W_2^t\,\mathbf{f_1}(W_1^t\mathbf{x_t + b_1)} + \mathbf{b_2}) + \mathbf{b_3} + \mathbf{\epsilon_t}
\end{eqnarray*}
\noindent where $W_1 \in \mathbb{R}^{(p+1)\times n_h}$, $W_2 \in \mathbb{R}^{n_h \times n_h}$ and $W_3 \in \mathbb{R}^{n_h\times n_o}$ are the weight matrices and $b_1 \in \mathbb{R}^{n_h}$, $b_2 \in \mathbb{R}^{n_h}$ and $b_3 \in \mathbb{R}^{n_o}$ are the bias vectors.
There are only two activation functions (vectorized $\mathbf{f_1}$ \& $\mathbf{f_2}$) for the hidden layers as the last layer uses Identity as the activation function. For the hidden layer nodes, during preliminary testing, we tried using multiple activation functions such as TanH, Sigmoid, and Exponential Linear Unit (eLU). However, we achieve the best performance with the Rectified Linear Unit (ReLU) activation function as specified by the NAS. Number of hidden neurons $n_h$ for both hidden layers is 1024 and the number of output layer neurons $n_o$ are 14 for regular time-series and 28 for the augmented time-series. While collecting out-of-sample forecasts for the augmented series, we collect only the even horizon forecast values because the odd horizons represent the half-a-day forecasts when compared to the original time-series. 

As neural networks are supervised learning models, they require training and testing data in the form of input matrix of size $m \times n$, where $m$ is the number of samples and $n$ is the number of features. The output/response may be a $m$ dimensional vector if we want to predict a single output per sample or a $m \times k$ dimensional matrix, where $k$ is the number of values we want to predict for each sample in the training/test set. In this study, we consider the direct multi-horizon forecasting method \cite{taieb2012recursive} for the neural networks and thus, have $k=14$ or $k=28$ to forecast all horizons directly. We have $n = (p+1)$ as we consider $p$ lags and the current time $t$. 

In order to generate the $m \times (p+1)$ dimensional input matrix and $m \times k$ dimensional output matrix, we use a sliding window approach \cite{edwards1997traffic} to convert the time-series into a supervised learning problem. Starting from the first value we take $p$ consecutive values in the time-series and along with the current time $t$, it forms a $(p+1)$ dimensional vector which is a single sample in the input matrix. Next, starting from $(p+1)^{th}$ timepoint, we take the next $k$ samples which forms the first $k$ dimensional vector in the output matrix. This approach is repeated till the end of the time-series is reached with a stride of 1 timepoint between each sample.

The loss function used is mean absolute percentage error (MAPE) and The optimizer used is Adaptive Momentum Estimate (ADAM) \cite{kingma2014adam}.

\subsection{Gated Recurrent Unit (GRU)-based Autoencoders}

Autoencoders are neural network-based models which may be several layers deep and generate compressed feature vectors in between the layers used for modeling the input data. The idea is that we use 1 or several modeling layers to convert the input data into a feature vector represented by a certain dimensionality which usually less than the input data to that layer. This compressed feature vector is now passed on to the next modeling layer which will act as a decoder to model the information that will be encoded into the feature vector by the encoding layer. This can, potentially, give us  rich feature representations of the data and allow for better modeling of the data. The compressed feature vector is also called as the compressed knowledge representation of the original input data. This also allows for dimensionality reduction as the constrained knowledge representation forces the autoencoders to have only the important features from the input data required to model the given data. The intermediate feature vectors are usually fixed-length and the its size is treated as tunable hyperparameter while specifying the architecture of such networks.

GRU-based autoencoders (GRU-AE) are autoencoder architectures which use GRUs as the modeling layers in the encoder and decoder stages of the network. This allows to generate the knowledge representation using the sequencial modeling capapabilities of the GRUs. Although autoencoders are most often used to reconstruct the original input, in this modeling study we use Fully Connected Layer added at the output of autoencoder to use such networks for modeling the time-series data. The results are competitive and comparable to the ConvLSTM we use in this study.

\subsection{Convolutional-LSTM}

Several studies have demonstrated the viability of the deep hybrid modeling, using convolutions inside LSTM cells, over a single model \cite{shi2015convolutional,sutskever2014sequence}.
This type of architecture enables the training process to benefit from the advantages of both models while typically achieving results better than individual CNN and LSTM.
CNN can be applied to time-series data since neighboring information is supposedly relevant for the analysis of the time-stamped data.
However, this relevancy is short-term due to the constraint by the size of convolutional kernels. 
On the contrary, LSTM can store and output information during the training process by capturing the long-term time dependency of input data, when used together they form the ConvLSTM architecture.
ConvLSTM replaces matrix multiplication with convolution operation at each gate in the LSTM cell, leading to capture informative features by convolution operations and provide the LSTM cell with sequenced data. This is the main distinction between a regular LSTM and a ConvLSTM. The equations remain the same apart from the fact that all weight matrix multiplications are converted into convolution operations as given in the equations for LSTM above.
By stacking multiple ConvLSTM layers and forming an encoding-forecasting structure, it is possible to build an end-to-end trainable model for spatio-temporal forecasting.

\section{Methodology}
We use forecasts from a statistical model as the basis for our data augmentation approach for the neural network models.
\subsection{Data Augmentation}
Neural Networks learn the output estimation function from the training data iteratively through the process of back-propagation. Typically, this process is carried out by one of the gradient-based optimization algorithms such as gradient descent or ADAM and requires a large number of data samples given a large number of trainable parameters in these networks. This puts us at a challenging spot when the number of trainable parameters/weights of the neural networks far exceeds the number of data samples that are available to us for training.

With time-series data, this is usually a common scenario and a point where we have to think about data augmentation approaches to aid the training of such networks. One of the major findings from the results of the M4 competition \cite{makridakis2018m4} was the superiority of a hybrid approach that utilizes both statistical and ML features \cite{MAKRIDAKIS202054}. Based on these findings, we combine the first horizon forecasts generated by the SARIMA$(1, 0, 0)_{\times}(3, 1, 1)_7$ model to build an augmented version of the time-series such that these forecasts serve as a basis to calculate the intermediate day or half-a-day time point values for our daily COVID-19 data. In order to generate these intermediate time points, the next day's forecasted value is averaged with the previous day's true data value. Thus, this interleaved time-series is double the size of the original time-series allowing for better training of the neural networks. Apart from data augmentation, this also allows for statistical feature information to be fed into the neural networks in order to facilitate better generalization performance and thus improving their performance on out-of-sample forecasts.

For generating the forecasts from the SARIMA$(1, 0, 0)_{\times}(3, 1, 1)_7$ model, it is trained on entire time-series and we collect in-sample forecasts for half the size of the original time-series. The latter half of the forecasts are collected by out-of-sample forecasts by retraining the model only on the first half of the time-series.

\subsection{AutoML}
Automated machine learning (AutoML) refers to the process of automating the application of machine learning to real-world problems. It has shown promise in advancing the use of machine learning on a variety of problem areas and datasets as indicated in a detailed study by Xin He, Kaiyong Zhao, Xiaowen Chu \cite{he2021automl}.  Specifically, in the area of deep learning the concept of Neural Architecture Search (NAS) has shown promise in finding better neural network architectures tailored to the specific learning problem and the dataset. Techniques such as Bayesian Optimization (BO) for Hyperparameters \cite{ wu2019hyperparameter} improve and automate the tuning of the neural network architectures, whereas NAS tries to search for the best network architecture using the BO as its base for every iteration.

The AutoML framework used is Auto-Keras \cite{jin2019auto}. NAS is carried out on the training set where we allow the last 10\% of the training set to be used as validation set while evaluating the model performance. Repeated trials help us find a baseline architecture suitable for the task which we further fine-tune by the inclusion of the validation set into the training set to generate the out-of-sample forecasts on the test set.

\subsection{Rolling Validation for Multiple Horizons}


Due to the dependencies of instances near each other
in time, $k$-fold cross-validation will not work for time-series data.
A simple form of rolling validation divides a dataset into an
initial training set and test set.
For example, in this work, the first 60\% of the samples is taken
as the training set and rest as the test set.
For horizon $h = 1$ forecasting, the first value in the test set
is forecasted based on the model produced by training on the training set.
The error is the difference between the actual value in the test set and
the forecasted value which is used to calculate the symmetric mean absolute percentage error (sMAPE).

For simple, efficient models, the process of rolling forward
to forecast the next value in the test set often involves retraining
the model by including the first value in the test set in the training set.
To maintain the same size for the training set, we
remove the first value from the training set.
We adjust the frequency of retraining for the statistical models such that we forecast $kt$ samples ahead in the test set before including them in the training set and retraining our model. In this study, the frequency of retraining $kt$ for the statistical family of models is set to 5 samples. The neural networks are only trained once on the training data without any retraining for generating the out-of-sample forecasts.

\section{Results}

Our empirical results for various statistical and neural network models are shown in Table \ref{tab:arima} and Table \ref{tab:nets}, respectively. We collect Multi-Horizon rolling forecasts on daily deaths for the next 2 weeks i.e. from $h=1$ through $h=14$ and use sMAPE as our primary performance metric which is one of the standard performance metrics in time-series forecasting \cite{makridakis2018m4, taieb2012recursive}.


\begin{small}
\begin{table}[!h]
\begin{center}
\caption{Multi-Horizon ($h$) Rolling Forecasts: Statistical Models (sMAPE).}
\label{tab:arima}
\begin{tabular}{|c|c|c|c|c|c|c|c|c|} \hline
Horizon   & RW & AR  & ARIMA & SARIMA  \\ \hline \hline
in-sample & 30.26  & 21.83 (13)  & 19.40 (19)  & \textbf{17.87 (12)} \\ \hline
$h = 1$   & 27.75  & 17.20 (8)   & 16.80 (12)  & \textbf{15.40 (1)} \\ \hline
$h = 2$   & 43.49  & 19.20 (10)  & 19.10 (7)   & \textbf{17.30 (1)} \\ \hline
$h = 3$   & 50.23  & 19.00 (10)  & 18.70 (9)   & \textbf{18.60 (4)} \\ \hline
$h = 4$   & 50.44  & 19.30 (10)  & \textbf{18.70 (9)}   & 18.80 (6)   \\ \hline
$h = 5$   & 44.61  & 19.80 (9)   & \textbf{19.10 (9)}   & 19.60 (1)   \\ \hline
$h = 6$   & 30.31  & 19.60 (9)   & \textbf{18.70 (9)}   & 19.20 (9)   \\ \hline
$h = 7$   & \textbf{17.36}  & 19.60 (12)  & 18.50 (8)   & 19.40 (9)   \\ \hline
$h = 8$   & 29.92  & 23.60 (15)  & \textbf{21.30 (12)}  & 22.10 (9)   \\ \hline
$h = 9$   & 42.49  & 27.20 (11)  & 24.80 (10)  & \textbf{24.60 (9)}   \\ \hline
$h = 10$  & 47.77  & 27.40 (15)  & 25.00 (14)  & \textbf{24.90 (8)}   \\ \hline
$h = 11$  & 49.36  & 27.70 (15)  & \textbf{24.10 (15)}  & 25.00 (9)   \\ \hline
$h = 12$  & 44.91  & 28.00 (15)  & \textbf{24.50 (15)}  & 25.10 (9)   \\ \hline
$h = 13$  & 32.30  & 28.50 (9)   & \textbf{24.70 (15)}  & 25.30 (9)   \\ \hline
$h = 14$  & \textbf{24.05}  & 29.30 (9)   & 24.10 (15)  & 25.40 (22)   \\ \hline
\end{tabular}
\end{center}
\end{table}
\end{small}

In order to obtain the best performing model for each horizon we train and evaluate several models of each type by changing the auto-regressive non-seasonal order ($p$) of such models. The ($p$) in Table \ref{tab:arima} indicates the auto-regressive non-seasonal order for the model with lowest sMAPE for the given horizon.

\begin{small}
\begin{table*}[h]
\begin{center}
\caption{Multi-Horizon ($h$) Rolling Forecasts: Neural Network models (sMAPE).}
\label{tab:nets}
\begin{tabular}{|c|c|c|c|c|c|c|c|c|} \hline
Horizon   & NN      & AUG-NN  & GRU-AE & AUG-GRU-AE & ConvLSTM & AUG-ConvLSTM  \\ \hline \hline
in-sample & 11.42	& \textbf{10.65}   & 13.93	& 11.32      & 15.04	& 11.99  \\ \hline
$h = 1$   & 15.90	& \textbf{14.36}   & 18.87	& 15.32      & 18.59	& 16.37  \\ \hline
$h = 2$   & 18.43	& \textbf{15.73}   & 19.48	& 17.37      & 17.88	& 17.38  \\ \hline
$h = 3$   & 18.62	& \textbf{16.79}   & 18.75	& 17.91      & 18.05	& 17.01  \\ \hline
$h = 4$   & 19.06	& 17.82   & 19.38	& 17.69      & 18.59	& \textbf{16.88}  \\ \hline
$h = 5$   & 23.45	& \textbf{18.43}   & 19.92	& 18.61      & 20.77	& 18.90  \\ \hline
$h = 6$   & 20.18	& \textbf{18.72}   & 21.08	& 18.81      & 22.32	& 19.65  \\ \hline
$h = 7$   & 22.77	& 19.35   & 22.18	& \textbf{18.06}      & 24.09	& 22.79  \\ \hline
$h = 8$   & 22.27	& 20.13   & 25.21	& \textbf{19.35}      & 26.02	& 23.81  \\ \hline
$h = 9$   & 24.83	& 21.15   & 24.29	& \textbf{20.63}      & 26.28	& 22.87  \\ \hline
$h = 10$  & 24.69	& 21.94   & 24.52	& \textbf{20.64}      & 26.80	& 22.40  \\ \hline
$h = 11$  & 23.13	& 22.57   & 23.77	& \textbf{20.84}      & 26.81	& 23.08  \\ \hline
$h = 12$  & 25.49	& 23.82   & 24.13	& \textbf{23.30}      & 28.44	& 25.20  \\ \hline
$h = 13$  & 26.47	& 24.46   & 25.25	& \textbf{23.06}      & 30.72	& 28.25  \\ \hline
$h = 14$  & 25.97	& 24.28   & 28.74	& \textbf{21.76}      & 32.96	& 30.49  \\ \hline
\end{tabular}
\end{center}
\end{table*}
\end{small}

For AR($p$) model, we evaluate $p=1$ through $p=22$. Similarly, for ARIMA$(p, 1, 0)$ model, we use a fixed single difference with no MA component for all horizons, whereas $p$ goes from 1 through 22. 
The SARIMA$(p, 0, 0)_{\times}(3, 1, 1)_7$ uses no differencing and no MA component, whereas it includes $P=3$ and $Q=1$ seasonal components with 1 seasonal differencing component and a seasonal period of 7 days.

The highlighted values in each row indicate the best performing model for the given horizon having the lowest sMAPE score amongst all the statistical models. The SARIMA model performs well overall, however the ARIMA model is competitive on several horizons, whereas the RW model having the lowest error for horizons 7 and 14.


Table \ref{tab:nets} compares the performance of two configurations of neural network-based models. The configuration I refers to the feed forward neural network, configuration II refers to the GRU-based Autoencoder and configuration III refers to the ConvLSTM network. NN column refers to the neural networks run on regular time-series with 376 samples, whereas AUG-NN refers to the Augmented-Neural-Network run on the data augmented time-series with 750 samples. Since the augmented time-series has a resolution of half-a-day, we collect the sMAPE for every even horizon output from $h=2$ through $h=28$ to gives us the actual 14 day ahead forecasts.

Due to the sliding window approach used to generate the supervised training data for neural networks, there is an offset in the number of samples in the testing set at the end of our time-series. This happens when the sliding window rolls over the end of the time-series as we have to accommodate for $h$ samples to perform multi-horizon forecasting. To generate the out-of-sample forecasts, we require several neural network models each having varying horizons. Since, this is computationally inefficient, we skip the last $h-1$ samples from the test set while calculating the performance metrics on the neural network-based models.

The data augmented networks performed much better compared to the non-augmented ones due to the increased amount of training samples as well as the information fed from the SARIMA model. The AUG-GRU-AE performed the best for longer horizons while the AUG-NN was the best performing model for shorter horizons.

\begin{figure}[htbp]
\centerline{\includegraphics[width=88mm]{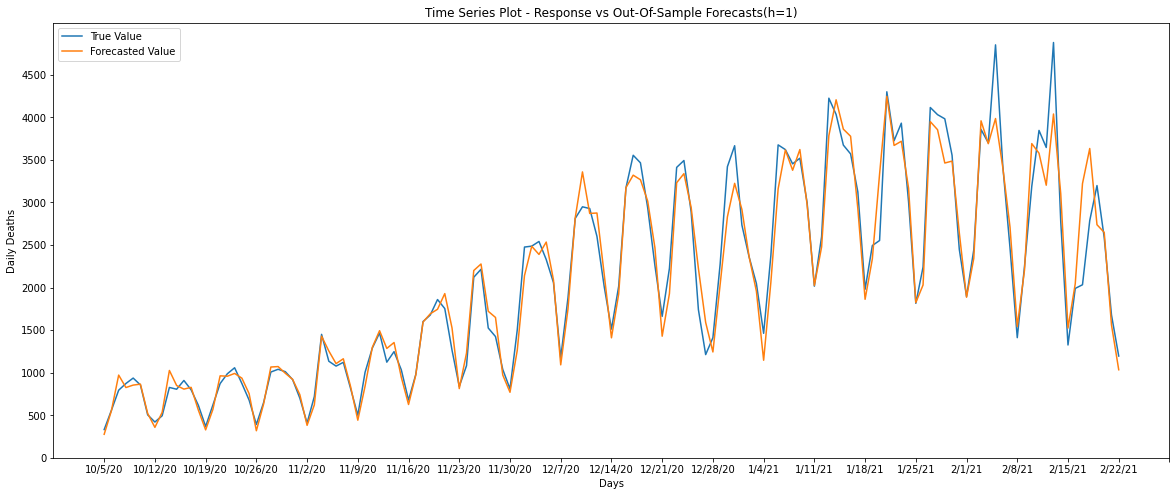}}
\caption{Out-Of-Sample Forecasts ($h=1$)}
\label{fig:outforecasts}
\end{figure}

Figure \ref{fig:outforecasts} shows the true values (in blue) vs. the horizon 1 out-of-sample forecasts (in orange) of the AUG-NN model.
The forecasts in the initial days closely follow the true death count and is only off in the later days where fluctuations and trend is higher.

\begin{small}
\begin{table}[!h]
\begin{center}
\caption{Improvement Due to Augmentation}
\label{tab:compare}
\begin{tabular}{|c|c|c|c|c|c|c|c|c|} \hline
Horizon   & SARIMA & NN & AUG-NN & Improvement (\%)  \\ \hline \hline
$h = 1$   & 15.40 & 15.90 & 14.36  & 9.69   \\ \hline
$h = 2$   & 17.30 & 18.43 & 15.73  & 14.65   \\ \hline
$h = 3$   & 18.60 & 18.62 & 16.79  & 9.83   \\ \hline
$h = 4$   & 18.80 & 19.06 & 17.82  & 6.51   \\ \hline
$h = 5$   & 19.60 & 23.45 & 18.43  & \textbf{21.41}   \\ \hline
$h = 6$   & 19.20 & 20.18 & 18.72  & 7.23   \\ \hline
$h = 7$   & 19.40 & 22.77 & 19.35  & 15.02   \\ \hline
$h = 8$   & 22.10 & 22.27 & 20.13  & 9.61   \\ \hline
$h = 9$   & 24.60 & 24.83 & 21.15  & 14.82    \\ \hline
$h = 10$  & 24.90 & 24.69 & 21.94  & 11.14    \\ \hline
$h = 11$  & 25.00 & 23.13 & 22.57  & 2.42   \\ \hline
$h = 12$  & 25.10 & 25.49 & 23.82  & 6.55  \\ \hline
$h = 13$  & 25.30 & 26.47 & 24.46  & 7.59  \\ \hline
$h = 14$  & 25.40 & 25.97 & 24.28  & 6.51   \\ \hline
\end{tabular}
\end{center}
\end{table}
\end{small}

Table \ref{tab:compare} shows the improvement in AUG-NN over the neural network trained on the original time-series. In terms of improvement due to augmentation, on average we see a 10.21\% improvement across all horizons with a maximum improvement of 21.41\% seen at horizon 5 forecasts. While the NN model does not perform better than the SARIMA model, the AUG-NN model substantially outperforms the SARIMA model. Also, the GRU-AE model saw an improvement of 12.97\% and the ConvLSTM model saw an improvement of 9.62\% on average across all horizons.

\section{Conclusions and Future Work}

In this paper, we present a data augmentation method in order to improve the quality of out-of-sample forecasts for neural networks used for time-series forecasting. Our empirical analysis shows that, in comparison to the traditional statistical models, neural networks struggle to produce better quality forecasts on intermediate length time-series forecasting where the sample size is limited.

However, with the data augmentation paired with AutoML to search for the optimal neural network architecture, the performance of such neural networks is significantly boosted making them better than the best performing statistical models. We also show that not just with the feed forward networks, this performance improvement is also possible on other deep learning models such as the GRU-AE and ConvLSTM model. 

When compared with the methods using hybrid approaches described in our related work, our method is much simpler to implement and offers relatively larger performance improvements. The simplicity of our data augmentation method makes it an easy yet powerful technique to improve the performance of neural networks on time-series data.

In the context of forecasting on intermediate time-series forecasting, our work serves as a baseline to improve the performance of neural networks and encourages further research work on this study to incorporate additional deep learning architectures along with the epidemiological models such as SEIR/SIR model for more in-depth analysis and comparison.

\nocite{box2015time}

\bibliographystyle{IEEEtran}
\bibliography{IEEEabrv,main}

\end{document}